\documentclass[10pt,twocolumn,letterpaper]{article}

\usepackage[pagenumbers]{cvpr} 

\usepackage[table]{xcolor}

\definecolor{cvprblue}{rgb}{0.21,0.49,0.74}
\usepackage[pagebackref,breaklinks,colorlinks,allcolors=cvprblue]{hyperref}
\usepackage{tikzducks}
\usepackage{bm}
\usepackage{multirow}
\usepackage{cuted}

\newcommand\blfootnote[1]{%
  \begingroup
  \renewcommand\thefootnote{}\footnote{#1}%
  \addtocounter{footnote}{-1}%
  \endgroup
}

\title{
Predicting 4D Hand Trajectory from Monocular Videos}

\newcommand*{\ours}[0]{HaPTIC\xspace}

\newcommand*{\wtf}[0]{Weak2Full\xspace}
\newcommand*{\wtfMid}[9]{Weak2Full\xspace}
\newcommand{\supmat}{{{Appendix}}\xspace}

\author{
Yufei Ye\textsuperscript{1} \qquad Yao Feng\textsuperscript{2} \qquad Omid Taheri\textsuperscript{2} \qquad Haiwen Feng\textsuperscript{2} \\ \qquad Shubham Tulsiani \textsuperscript{1*} \qquad Michael J. Black \textsuperscript{2*}  \\
\textsuperscript{1}Carnegie Mellon University  \qquad \textsuperscript{2} Max Planck Institute for Intelligent Systems \\
{\tt \small \href{https://judyye.github.io/4dhands}{https://judyye.github.io/haptic-www}}
}

\begin{document}
\maketitle

\begin{strip}\centering
\vspace{-1.5cm}
\includegraphics[width=\textwidth]{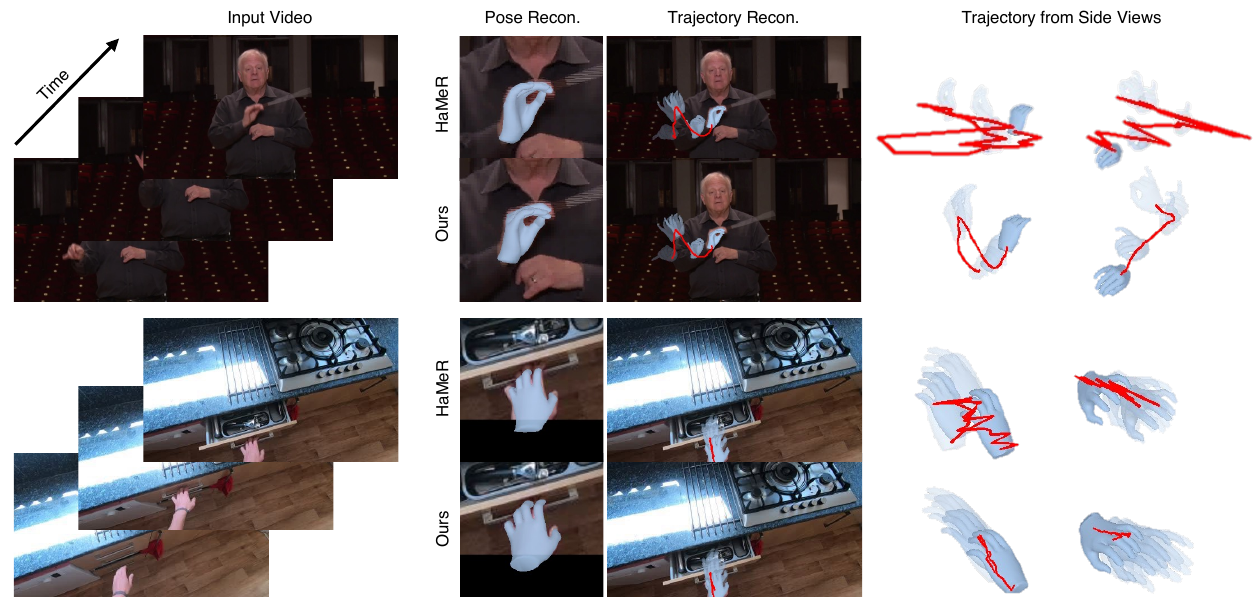}
\vspace{-0.7cm}
\captionof{figure}{Given a monocular video depicting hand motion, \ours~reconstructs both \textbf{Ha}nd \textbf{P}ose (Pose Recon.) and 4D hand \textbf{T}rajectory \textbf{i}n consistent global \textbf{C}oordinate (last 3 columns). The existing method produces convincing reprojection but its 4D trajectory is not plausible (side view). In contrast, our method can generate faithful 4D trajectories. The opaque hand shows reconstruction of the last frame while semi-transparent hands visualize reconstructions from previous frames. Red curves visualize root trajectories. 
\label{fig:teaser}}
\end{strip}

\begin{abstract}
We present \ours, an approach that infers coherent 4D hand trajectories from monocular videos. 
Current video-based hand pose reconstruction methods primarily focus on improving frame-wise 3D pose using adjacent frames rather than studying consistent 4D hand trajectories in space. Despite the additional temporal cues, they generally underperform compared to image-based methods due to the scarcity of annotated video data. 
To address these issues, we repurpose a state-of-the-art image-based transformer to take in multiple frames and directly predict a coherent trajectory. 
We introduce two types of lightweight attention layers: cross-view self-attention to fuse temporal information, and global cross-attention to bring in larger spatial context.
Our method
infers 
4D hand trajectories 
similar to the ground truth
while maintaining strong 2D reprojection alignment. We apply the method to both egocentric and allocentric videos. It significantly outperforms existing methods in global trajectory accuracy while being comparable to the state-of-the-art in single-image pose estimation.
\end{abstract}  \blfootnote{$^*$ Equal advice. }  

\section{Introduction}

Consider the video frames shown in Figure \ref{fig:teaser}. One can infer not only the articulation of the hand (3D pose) depicted in every frame but also understand the hand motions across frames and in space. For example, in the bottom row of Figure~\ref{fig:teaser}, we understand that the hand extends towards the drawer and then retracts along a nearly identical path.
The ability to infer 4D hand trajectory (3D space plus time) is important for many downstream tasks, such as reasoning about hand-object interactions in a global scene \cite{ma2022hand}, AR/VR applications \cite{pei2022hand}, and imitation learning in robotics \cite{bharadhwaj2024towards, wang2024dexcap}.
Unfortunately, despite impressive progress in hand pose estimation that infers frame-wise 3D pose, current methods still struggle to coherently put the hands into a global 3D space. In this work, we address this problem with a system that can infer coherent 4D hand trajectory and pose from monocular videos.

While no prior work is dedicated to direct 4D hand trajectory prediction, common practices often involve first predicting per-frame 3D hand poses, ‘lifting’ them to a world coordinate system, followed by test-time optimization \cite{hasson2021towards,duran2024hmp,patel2022learning}. 
The de-facto lift method uses a weak-to-full perspective transformation (\wtf)~\cite{hamer,frankmocap}, which places the predicted hand at a certain distance given the camera intrinsics and predicted scale (Sec.~\ref{sec:wtf}). However, we find that this operation introduces a large error in 4D motion, and that even post-processing optimization struggles to correct it. An alternative lift approach uses estimated depth \cite{zoe,epic_fields} but occlusion of the hand from object or scene interaction, or from the other hand, makes the induced trajectory inaccurate. 
One alternative philosophy is to consider the more holistic task of full body estimation in 4D and then the hands are defined relative to the body \cite{wham,yi2024estimating}.
Unfortunately, this requires the full body to be largely visible which is typically not true for videos focusing on hand manipulation and egocentric video.

In contrast, we formulate the problem as a 4D inference problem from video and leverage an implicit data-driven prior. 
Given a video sequence as input, we output MANO  \cite{mano} hand parameters with the wrist in global coordinates.
Unlike hand-designed lifting, our approach better captures prior information about hand motion and does not rely on explicit full-body estimation.
This results in, to our knowledge, the first feed-forward method that estimates consistent 4D hand trajectories directly from monocular video. 

\begin{figure*}
    \centering
    \includegraphics[width=\linewidth]{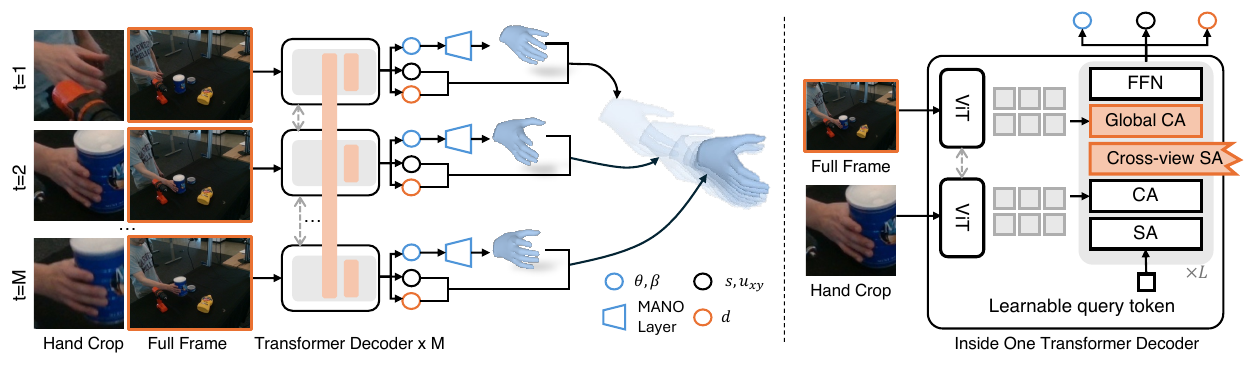}
    \vspace{-1.8em}
    \caption{\textbf{Overall pipeline (left)}:  \ours~ extends image-based model HaMeR. \ours~takes in $M$ frames at a time and passes them through image towers that share weights. Each image tower outputs  MANO parameters in local coordinate, and trajectory parameters $d,u_{xy}$  that directly places the predicted local hand to global 4D trajectory. 
    \textbf{Inside one image tower (right)}: The image tower is based on transformer decoder. For each block, we add a cross-view self-attention layer (Cross-view SA) to fuse temporal information from other frames and a cross-attention (Global CA) to features of the original frames. Orange indicates new components introduced by ours compared to HaMeR. }
    \label{fig:overview}
    \vspace{-0.5em}
\end{figure*}

The problem, however, is the lack of training data containing video paired with 3D hand annotations in global coordinates.
While some such data exists, it is much scarcer and less diverse than image-based training data.
To address this limitation, our key idea is to heavily leverage the single-image data in terms of both models and training data.
We then incorporate video training data in such a way that we need significantly less data than if we trained the 4D inference with video alone.
Specifically,  we repurpose the state-of-the-art image-based hand pose estimator \cite{hamer} to incorporate video inference by injecting two types of lightweight attention layers (Fig.~\ref{fig:overview}). 
The first is a cross-view self-attention layer that sees across multiple frames to leverage temporal cues. 
The second is a global-context cross-attention layer that sees the original frames (from which the input hand images are cropped) to gain a larger field-of-view of the scene. 
By design, the network allows both video and images as input –- when the input is an image, the cross-view attention simply degrades to attending to itself. 
This allows us to intersperse both small-scale video data and large-scale image data in training batches.
In this way, our new model, called \ours, maintains the robustness and generalization ability of single-image methods but enriches them with temporal information from video.

We evaluate \ours on both allocentric and egocentric videos. 
We analyze the drawbacks of the prevalent practices, including when and why they fail. 
\ours significantly outperforms all other potential candidates in terms of global trajectory accuracy. It also provides a better initialization for test-time optimization. 
We carefully ablate our method to analyze our design choices. When \ours serves as an image-based pose estimator, we find that it even outperforms the state-of-the-art image-based method in terms of per-frame hand pose 2D alignment.  
Finally, we show the generalization ability to in-the-wild videos and images by presenting more qualitative results. Our model will be publicly released for broader use and further research.

\section{Related Work}

\noindent\textbf{Image-Based Hand Pose Estimation.}
There is a rich body of work tackling hand pose estimation from image input and typical approaches can be categorized into template-based methods ~\cite{zhang2019HAMR, frankmocap} and template-free methods~\cite{tang2021handAR, moon2020i2l, metro,meshgraphormer,simplehand,hhmr}. Both benefit from diverse large-scale annotated data~\cite{hamer,interhand,cocow,mtc,freihand} and foundational backbones~\cite{resnet,vit,mamba}. 
While these methods perform effectively on single image frames, their predictions are all in the canonical local frame of the hand. Their predictions can be reprojected back to each video frame but do not induce coherent 4D trajectories in space. 
Yet, we believe that these models can establish a solid foundation for video-based hand estimation. Specifically, our work is built upon the state-of-the-art image model, HaMeR~\cite{hamer}, which is a large transformer trained on large-scale datasets to predict the parameters of a 3D hand model (MANO~\cite{mano}).

\smallskip\noindent\textbf{Video-Based Hand Pose Estimation.}
Hand tracking in 3D space has a long history and is central to many applications. To get the most reliable tracking, methods rely on additional sensors like IMUs, magnetic sensors, or specialized hardware  \cite{wang2009real,projectaria,caeiro2021systematic,garcia2018first}. 
Multi-camera video systems can also provide pseudo ground truth \cite{ruocheng_piano,h2o,arctic,interhand}. 
While recent approaches reconstruct hand motion from monocular videos \cite{semihandobj,liu2022hoi4d,ye2023ghop,hasson2021towards,patel2022learning}, these methods are based on test-time optimization using a  learned prior, which is time-consuming; they can take a couple of minutes to hours to process clip of a few seconds.
There are a few feed-forward video-based methods \cite{yang2020seqhand,multiviewvideohpe,arctic,fu2023deformer,omnihands} but these are not as robust or general as the single-image methods due to a lack of annotated video training data. 
More importantly, they focus on improving local pose and do not tackle the problem of global trajectory estimation.  Our approach is the first feed-forward method to reconstruct the global trajectory from monocular videos.

\smallskip\noindent\textbf{4D Whole-Body Reconstruction.} 4D whole-body trajectory
estimation in global space has seen recent attention, either using feed-forward predictions \cite{glamr,trace, Li2022DD, Yu2021Human,wham,Pavlakos2022Human}, or global optimization \cite{pace,ye2023decoupling,Sun2021Monocular,Gartner2022Trajectory} that decouples human motion from camera motion ~\cite{Schonberger2016SfM} using learned human motion prior~\cite{Rempe2021HuMoR}.  
These methods could potentially place hands in global coordinates by explicitly attaching the hand pose to the wrist. 
However, this requires the full body to be largely visible, which is often not the case in videos focusing on hand manipulation and egocentric videos. 
Instead of leveraging whole body pose as a form of explicit global context, the global spatial cross-attention in \ours provides the implicit context of the scenes and humans. 
We demonstrate that even when the full body is visible, our method outperforms the state-of-the-art method~\cite{wham} in this category.

\smallskip\noindent\textbf{Adapting Image-Based Models to Videos.} We are inspired by the idea in generative AI that upgrades pretrained single-image models~\cite{Karras2019stylegan2,Rombach2021HighResolutionIS,ding2022cogview2} to other modalities through lightweight adaptation. Image-based models are upgraded to video generative models via multi-frame hierarchical training~\cite{hong2022cogvideo}, spatiotemporal factorization~\cite{Singer2022MakeAVideoTG}, video noise priors~\cite{Ge2023PreserveYO}, and temporal attention combined with 3D convolutional blocks~\cite{Blattmann2023AlignYL,Blattmann2023StableVD}.
Image models are similarly extended to 3D generation~\cite{Shi2023MVDreamMD,Liu2023SyncDreamerGM,Gao2024CAT3DCA,hu2024mvd} by introducing multi-view consistency or geometry-guided attention mechanism. In our work, we repurpose a pretrained single-image hand estimator to take in multiple frames. Importantly, this allows us to train on both video and image datasets.

\section{Method}
Given a monocular video with detected hand bounding boxes $\{I_t, B_t\}_{t=1}^T$, \ours infers hand articulations and their 3D locations in the camera coordinate system in a feed-forward manner.  
We first revisit the preliminary parametric hand model MANO and analyze why the de-facto approach for uplifting is undesirable (Sec.~\ref{sec:wtf}). 
We then introduce our simple yet effective parameterization of hand trajectory (Sec.~\ref{sec:uplift}). 
Finally, we describe our network architecture that predicts hand trajectories and explain our design choices, 
which address the problem of limited video training data (Sec.~\ref{sec:network} and ~\ref{sec:training}).

\subsection{Preliminaries: MANO and \wtf}
\label{sec:wtf}

Like prior work~\cite{hamer,frankmocap,fu2023deformer}, we use the parametric hand model, MANO~\cite{mano}.
MANO takes a 48-dim pose parameter vector $\bm \theta$, including finger articulation and wrist orientation, as well as 10-dim shape parameter vector $\bm \beta$ and it outputs a hand mesh surface $\mathcal M^l (\bm{\theta}, \bm{\beta})$ and hand joints $\bm J^l(\bm \theta, \bm \beta)$. Superscripts, $l$, denote coordinates in the local MANO frame. 

\smallskip\noindent\textbf{Weak-to-Full Perspective.} In order to overlay the predicted (local) hand mesh onto the input image crop, prior work also predicts a weak-perspective camera, which 
projects a 3D point by scaling and translating the $xy$ components, $\bm X^i_{x, y} = s\bm X^l_{x,y} + u_{x,y}$, where superscript $i$ denotes coordinate in the image frame. 
Note that although this results in 2D image alignment, a weak-perspective camera does not give the 3D position in the camera space  $\bm X^c$. To obtain a 3D position, the widely adopted approach is to use weak-to-full perspective transformation (\wtf)~\cite{hamer,cliff,frankmocap,hmr2,hasson2021towards,kanazawa2019learning}. 
It uses an assumed focal length $f$ as input and places the hand at a distance such that the re-projection of the transformed hand is approximately at the predicted scale under that assumed perspective cameras $\bm X^c = \bm X^l + (u_x, u_y, f/s)$.  

\begin{figure}
    \centering
    \includegraphics[width=\linewidth]{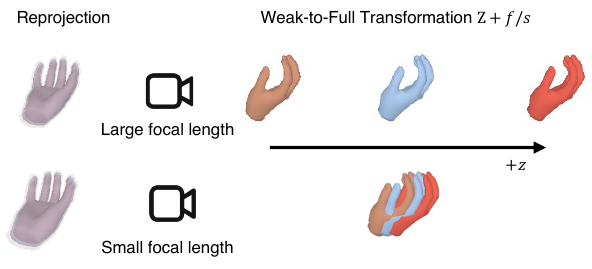}
    \caption{\textbf{Toy example of \wtf~ transformation: } The predicted scale $s$ of both yellow and red hands are only 6\% off the blue hand. Their reprojections appear similar yet the 3D position induced by \wtf~varies a lot with large focal length.}
    \label{fig:weak2full}
    \vspace{-1.5em}
\end{figure}

\smallskip\noindent\textbf{What's wrong with Weak-to-Full?} 
Looking closely at the offset in the $Z$-axis, $f / s$, in the \wtf process, we notice that the error in the predicted scale $s$ affects the 3D location (4D trajectory for videos). 
As shown in Fig.~\ref{fig:weak2full}, the trajectory is more prone to error in the predicted scale when the focal length is larger. 
Furthermore, even with optimal scales, different estimates of the focal length induce different motions in 3D space.

\subsection{Parameterize 4D Hand Trajectory}
\label{sec:uplift}
In contrast with the \wtf transformation that parameterizes hand trajectory with scale and focal length, \ie $\text{\wtf}(\{u_{x, t}, u_{y, t}, s_t, f_t\}_{t=1}^T$), we advocate for directly representing hand trajectory in the metric space, which is actually the most natural motion parameterization.  
Specifically, we simply predict \textit{change of depth} relative to the first frame: $ \bm X^c = \bm X^l + (u_x, u_y, \Delta d + d_1)$. 
$d_1$ is the estimated offset from the first frame, either from the \wtf transformation on the first frame or from depth prediction. We predict metric depth \textit{change}, which is invariant to the distance to the camera, instead of absolute metric depth because of depth ambiguity -- it is hard to tell if the hands are $5m$ away or $5.5m$ away but easier to infer that hand is moving by $5cm$ between frames. We learn a network to predict these parameters $\Delta d_t, u_{xy,t}$ along with  $ \bm \theta_t$ directly from video input.  


\subsection{Repurpose Image-Based Model for Videos} 
\label{sec:network}
Training a generalized video model from scratch requires a lot of data.  
Due to the lack of 3D annotated video data, we choose to extend the state-of-the-art hand pose image-based model (HaMeR) and design our network architecture to best leverage the existing model. We then finetune its weights to repurpose it into a video model.

\smallskip\noindent\textbf{Multi-Frame Siamese Network. } \ours, like HaMeR, is also a transformer-based model. 
Instead of taking a hand crop from a single image, we modify the network to take in $M$ crops at a time, along with their full frames (Fig.~\ref{fig:overview} left). Each frame is processed in a Siamese-network manner (sharing weights). The network is adapted from HaMeR with two types of lightweight adaption layers to fuse information among frames. 
In addition to the original prediction heads to output MANO parameters $\bm \theta, \bm \beta$ and weak-perspective cameras parameters $s, u_{x,y}$, we introduce an additional prediction head to predict metric depth change in order to uplift local hands to global hand trajectory.

\smallskip\noindent\textbf{Light-Weight Adaption Layers. } Each frame (and crop) is first passed to a ViT backbone, followed by a transformer decoder. 
As shown in Fig.~\ref{fig:overview} right, 
inside each block of the transformer decoder, we insert a cross-view self-attention (SA) layer and a global cross-attention (CA) layer after the original CA layer. The cross-view SA layer fuses temporal information from other frames. We add a positional encoding of the frame number in this layer. 
Since hand crops alone discard global information, the global cross-attention layer takes in a larger-view context by attending to the full-frame ViT features. 
The overall architecture only adds 4\% new parameters.
We initialize the new attention layers to output zero such that the new layers do not destroy the pretrained structures at the start of training. 

\smallskip\noindent\textbf{Generalize to Image Reconstruction. } While \ours~ is designed for video reconstruction, it can naturally incorporate single images as input without any modification. In such a case, cross-view SA only attends to its own frame. The image-mode inference can simply be achieved by a reshape operation. 

\subsection{Learn with Interspersed Video and Image Data}
\label{sec:training}
We believe that generalization ability of video models can be achieved by training on diverse {\em image} datasets with only limited video data. 
For each training batch, half of the batch comes from video datasets and is trained in video mode, while the other half comes from image datasets and is trained in image inference mode.

\definecolor{first}{rgb}{1.0, .83, 0.3}
\definecolor{second}{rgb}{1.0, 0.93, 0.7}
\def \first {\cellcolor{first}}
\def \second {\cellcolor{second}}
\def \third {}

\begin{table*}[t]
\footnotesize
\begin{center}
\vspace{-1em}

\setlength{\tabcolsep}{4pt}
\resizebox{\linewidth}{!}{
\begin{tabular}{l l c c c c c c c c c c c }
\toprule
\multirow{2}{*}{Local Pose}& \multirow{2}{*}{Uplift} & \multicolumn{3}{c}{ARCTIC-EXO} 
& \multicolumn{3}{c}{DexYCB} & \multicolumn{3}{c}{ARCTIC-EGO} \\
\cmidrule(r){3-5} \cmidrule(r){6-8} \cmidrule(r){9-11}
 & & GA-MPJPE & FA-MPJPE & ACC-NORM  & GA-MPJPE & FA-MPJPE & ACC-NORM & GA-MPJPE & FA-MPJPE & ACC-NORM   \\
\midrule
HaMeR& \wtf & 7.88 & 21.79 & 81.19 & 3.28 & 12.66 & 14.39 & 2.91 & 5.53 & 11.95 \\
\midrule
HaMeR$\dagger$  & \wtf & 5.53 & 13.29 & 30.09 & 1.93 & 9.78 & 4.42 & 3.48 & 9.23 & 9.62 \\
HaMeR$\dagger$  & \wtf-GT & 6.05 & 15.13 & 36.59 & \second 1.84 & \second 9.48 & 4.27 & \first \textbf{1.84} & \first \textbf{3.66} & 4.39 \\
HaMeR$\dagger$  & ZoeDepth & \second 3.28 & \second 10.09 & \first \textbf{1.83} & 4.93 & 20.51 & \first \textbf{1.32} & 2.95 & 5.60 & \first \textbf{1.57} \\
HaMeR$\dagger$  & WHAM & 4.79 & 14.41 & 2.43 & -- & -- & -- & -- & -- & -- \\
\multicolumn{2}{c}{\ours$\dagger$ (Ours) } & \first \textbf{2.07} & \first \textbf{5.18} & \second 2.02 & \first \textbf{1.61} & \first \textbf{6.58} & \second 1.82 & \second 2.27 & \second 4.31 & \second 1.58 \\
\midrule
\multicolumn{2}{c}{\ours (Ours)} & {1.88} & {4.49} & {1.48} & 1.80 & {5.80} & 1.70 & 2.39 & 4.49 & {1.34} \\
\bottomrule
\end{tabular}
}
\caption{\textbf{Comparison with baselines.} We compare our method with different uplifting baselines on two allocentric datasets (ARCTIC-EXO and DexYCB) and one egocentric dataset (ARCTIC-EGO). We report trajectory errors in GA-MPJPE, FA-MPJPE, and ACC-NORM. 
$\dagger$ denotes dataset-specific finetuned models on top of the publicly released HaMeR on the same amount of training data under the same training schedule. Among them, dark yellow marks the best results and light yellow marks the 2nd best ones. 
\ours~ in the last row trains on a combination of datasets with a longer schedule (see Effect of Scaling paragraph).} 
\label{tab:compare}
\end{center}
\vspace{-2em}
\end{table*}

\smallskip\noindent\textbf{Losses.}
We leverage the best practices for parametric hand pose estimation from images and supervise the model with a combination of 4D (video), 3D, and 2D losses. 

We supervise the predicted hand trajectories of length $M$ in a global space with the ground truth joints in global coordinates: $
\mathcal L_{\text{4D}} = \sum_{t=1}^M \|\bm J^c_t - \hat {\bm J}^c_t \|_1.
$ 
Note that this global supervision sounds demanding but global coordinates are actually present in most multi-view capture datasets. 
We align joint trajectories by the root joint in the initial frame.

The rest of the losses follow the recipe of training image-based hand pose reconstruction.  
We directly supervise hand pose in the local frame in 3D space, including consistency of MANO parameters and 3D hand joints:
$
\mathcal{L}_{\text{3D}} = \|\bm J^l - \hat{\bm J}^l \|_1 + \|\bm \theta - \hat {\bm \theta} \|_2^2 + \|\bm \beta - \hat{\bm \beta}\|_2^2
$.
We also supervise the network with 2D keypoint reprojection loss:
$
\mathcal{L}_{\text{2D}} = \|\bm j - \hat{\bm j} \|_1 $. Lastly, 2D keypoint consistency alone may lead to unrealistic 3D poses. To encourage the generated hands to look realistic, we use an adversarial loss following prior work \cite{hamer,hmr2}: $\mathcal L_{adv} = (D(\bm \theta, \bm \beta) - 1)^2$.

\begin{figure*}
    \centering
    \includegraphics[width=\linewidth]{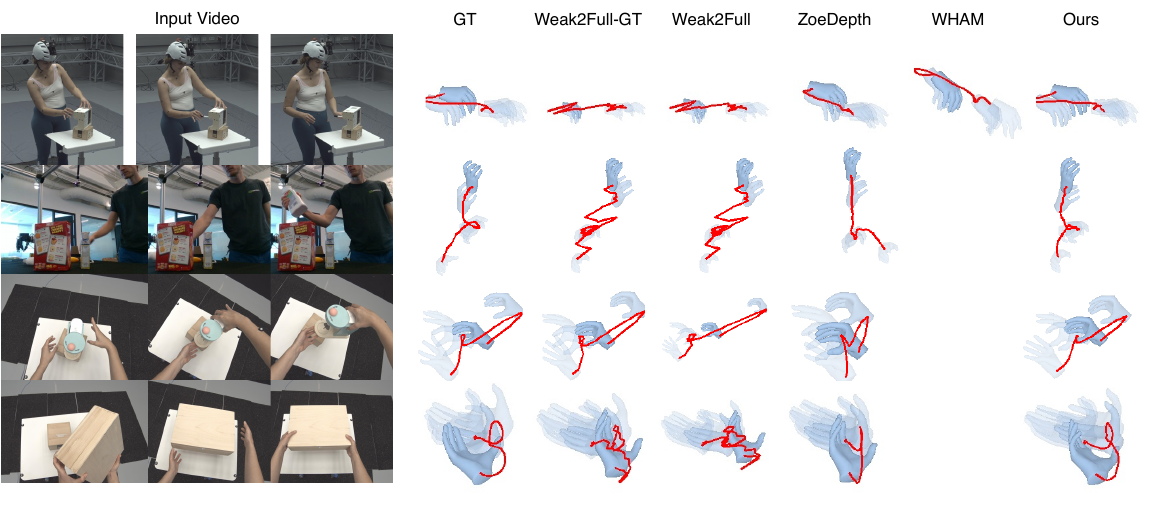}
    \vspace{-2.5em}
    \caption{\textbf{Qualitative comparison.}  
    We compare our approach qualitatively with other feed-forward baselines on all of three datasets. We show the first, middle, and last frames of an input video. We visualize hand root trajectory as red curves and five poses along time in a global coordinate viewed from the side. Poses from previous frames are visualized with transparency while the last frame reconstructions are opaque. We only visualize the right hand for clarity but all methods can reconstruct both the left and right hand. 
    We encourage readers to see results in videos on our website.   }
    \vspace{-0.5em}
    \label{fig:compare}
\end{figure*}
\section{Experiments}
We first evaluate our model on two allocentric datasets~\cite{arctic,dexycb}  and one egocentric dataset~\cite{arctic} in Sec.~\ref{sec:compare}. We compare with other feed-forward candidates and test-time optimization methods under a controlled training schedule and training data. 
Then, we scale up our model by training on a combination of 5 video datasets and 10 image datasets and evaluate performance of the unified model in terms of 4D trajectory and 3D poses (Sec.~\ref{sec:image-pose}). 
In Sec.~\ref{sec:ablation}, we analyze the effect of our network design.  Qualitative results on in-the-wild results from both videos and images are further shown in Sec.~\ref{sec:itw} and the \textbf{Suplemental Video}.

\smallskip\noindent\textbf{Training Data and Setup.}
We first compare dataset-specific models (ours and baseline) that are fine-tuned under controlled setting on DexYCB, or ARCTIC-EXO/EGO and test on their own split. 
\begin{itemize}
    \item DexYCB~\cite{dexycb} is an allocentric video dataset captured by multiple calibrated cameras. It features humans handling rigid objects on cluttered tables. Each sequence is 2-3 seconds long. We use the standard s0 split. 
    \item ARCTIC~\cite{arctic} provides both allocentric (-EXO) and head-mounted (-EGO) views of bi-manual manipulation of  articulated objects. Each sequence is $\sim$30 sec. We evaluate on the official subject-wise test split.   
    \item Combination of image data:  Our auxiliary image dataset is the same as the HaMeR training set, which consists of 10 datasets~\cite{freihand,interhand,mtc,rhd,cocow,halpe,mpiinzsl,ho3d,h2o,dexycb}.  
    \item Combination of video data: 
    Our general model is trained on a total of 5 video datasets. This includes the 4 datasets from the auxiliary image dataset that have 3D world coordinate annotations \cite{h2o,ho3d,interhand,dexycb}, as well as the ARCTIC dataset. The previous four are all allocentric, while only ARCTIC also contains egocentric views.
\end{itemize}
We set the window size of our model to 8, \ie $M=8$. During training, we augment the frame rate up to 6 fps to accommodate different speeds of hand motion. At test time, we use a sliding window with a 1-frame overlap to process long videos. More implementation details are in \supmat

\smallskip\noindent\textbf{Baselines.}
While there is a plethora of work to estimate per-frame (local) hand poses, there are no direct baselines that reconstruct 4D hand trajectories in a feed-forward manner. 
Therefore, we compare with common practices and possible alternatives to uplift local poses to 4D. The local hand poses for all baselines are from the same state-of-the-art image-based hand pose estimator~\cite{hamer} since local pose is not the main focus of our paper.  For a fair comparison, we finetune the released HaMeR on each dataset respectively with the same training schedule, denoted as $\dagger$.  
\begin{itemize}
    \item We first compare with the most commonly used uplift method, \wtf, with unknown camera intrinsics, which we set to the diagonal length of the image \cite{cliff,wham}. We also evaluate \wtf with ground truth intrinsics (\wtf-GT).  
    \item Metric depth computed using a monocular depth estimator can also be used to uplift local hand pose. We use ZoeDepth~\cite{zoe} to predict a pixel-wise metric depth map and solve for the optimal 3D offset that align it with the predicted hand surface from the local hand pose. 
    \item WHAM~\cite{wham} estimates body pose in global coordinates without hand pose. We use WHAM's estimated wrist orientation and location to uplift estimated local hand articulation.  Since WHAM requires most of the body  to be visible, this method only applies to ARCTIC-EXO. 
\end{itemize}

\smallskip\noindent\textbf{Evaluation Metrics.} We use metrics from whole-body pose estimation \cite{wham,glamr,slahmr} to evaluate 4D pose trajectory. In contrast to aligning poses per frame, GA-MPJPE globally aligns hand joints at all frames within one sequence before calculating its Mean Per Joint Position Error  (MPJPE). FA-MPJPE aligns the whole hand pose trajectory only using the pose in the first frame. ACC-NORM computes acceleration error compared with ground truth for each joint. 
The first two metrics evaluate trajectory quality globally while ACC-NORM focuses on local trajectory from adjacent frames. See \supmat for evaluation details. 

\subsection{4D Trajectory Reconstruction}
\label{sec:compare}
We report quantitative comparisons on three datasets in Table~\ref{tab:compare} and visualize the results in Fig.~\ref{fig:compare}. 

While finetuning the general HaMeR model on each dataset improves its own performance, the \wtf transformation produces significant jitter
in global space. This happens even when ground truth intrinsics are given. 
Our results highlight that trajectories from \wtf are sensitive to camera intrinsics. 
The error becomes even bigger with a large focal length (ARCTIC-EXO). 
 \wtf only performs well in terms of global alignment of trajectory (GA/FA-MPJPE) under the egocentric setting with ground truth intrinsics after finetuning. This is because the egocentric camera has a smaller focal length and, when hands are closer to the camera, it is reasonable to estimate hand distance by using its scale. 
 However, when the intrinsics deviate from ground truth, the trajectory induced from \wtf degrades.  

Uplifting by metric depth prediction gives a smoother trajectory (lower acceleration error). However, it does not perform well with  occlusions, which are present in the more cluttered scenes in DexYCB or due to hand-object interactions on ARCTIC-EGO. 

Global human body poses from WHAM only work when the person is largely visible, thus this approach cannot be used on DexYCB or egocentric videos. On ARCTIC-EXO, it still does not perform well because the method is optimized to predict the body's root trajectory instead of trajectories of distal extremities. 

In contrast, while \ours is able to predict equally good 2D alignment (Fig.~\ref{fig:teaser}, \ref{fig:result} and \supmat), Fig.~\ref{fig:compare} shows that our global trajectories are more consistent with ground truth. 
Quantitatively, \ours is among the top two methods across all metrics on all datasets. 


\definecolor{first}{rgb}{1.0, .83, 0.3}
\definecolor{second}{rgb}{1.0, 0.93, 0.7}
\def \first {\cellcolor{first}}
\def \second {\cellcolor{second}}
\def \third {}

\begin{table}[t]
\footnotesize
\begin{center}

\label{tab:arctic_exo}
\setlength{\tabcolsep}{2pt}
\resizebox{\linewidth}{!}{
\begin{tabular}{l c c c c c c }
\toprule
 & GA-MPJPE & FA-MPJPE & ACC-NORM & PA-MPJPE 
 \\
\midrule
\wtf-GT + Opt. & 4.38 & 10.90 & 1.14 & \second 0.54 \\
ZoeDepth + Opt. & \second 3.75 & \second 9.16 & \second 0.92 & 0.68 \\
\ours(Ours) + Opt. & \first \textbf{2.28} & \first \textbf{5.69} & \first \textbf{0.90} & \first \textbf{0.51} \\
\bottomrule
\end{tabular}
}
\caption{\textbf{Comparison with test-time optimization.} We compare trajectories after optimization when they are initialized from different baselines on the ARCTIC-EXO dataset. }
\label{tab:opt}
\vspace{-1.5em}
\end{center}
\end{table}

\begin{figure}
    \centering
    \includegraphics[width=\linewidth]{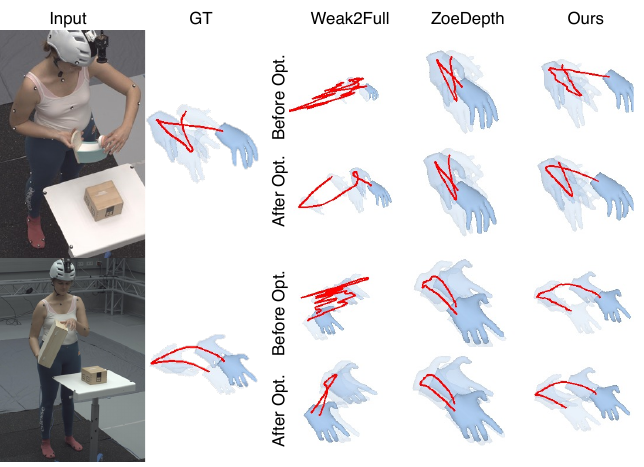}
    \vspace{-1.5em}
    \caption{\textbf{Qualitative comparison of optimization}: We compare results before and after optimization with trajectories initialized from baselines and \ours. The optimized trajectories are smoother but optimization struggles to correct global error. }
    \label{fig:compare_opt}
    \vspace{-0.5em}    
\end{figure}
\smallskip\noindent\textbf{Effect of Scaling.}  When trained on a combination of five video datasets, \ours even performs better than dataset-specific models (\ours$\dagger$) on allocentric views and comparably on egocentric data. 
The observation that more diverse data improves performance on in-domain test split is consistent with image-based pose estimation~\cite{frankmocap,hamer,hmr2}.
Since all video datasets~\cite{interhand,ho3d,h2o,dexycb} are all 3rd-person views except for ARCTIC,  we speculate that our model may further benefit from more egocentric data. 

\smallskip\noindent\textbf{Test-Time Optimization.}
Can jittery trajectories from feed-forward methods be improved by test-time optimization?  We further optimize the predicted trajectories for 1000 iterations with respect to 1) reprojection error to ground truth 2D keypoints, 2) acceleration, and 3) differences to the original prediction.  
Table \ref{tab:opt} and Fig.~\ref{fig:compare_opt} show that it is possible to make the predicted trajectory smoother but is much harder to correct global motions. In comparison, the motion prior learned by \ours~provides a better initialization for test-time optimization. PA-MPJPE stands for per-frame Procrustes-Aligned pose error.


\subsection{Image-based Estimation}
\label{sec:image-pose}
While \ours~is designed for 4D hand trajectories from videos, it generalizes to predict 3D hand poses from single images when video input is not available. Table~\ref{tab:image-pose} reports PCK at different thresholds on the HInt~\cite{hamer} benchmark. They are out-of-distribution datasets. The single-image hand predictions from \ours~outperform HaMeR and other baselines, whose margin is even more significant on occlusion subsplit (see \supmat). It is probably because the global CA layers are trained to capture full-frame context to infer hand poses, allowing plausible estimates even when the hand is occluded or blurry. We also report comparable 3D hand pose performance on other image-based benchmark~\cite{ho3d} in \supmat.

\definecolor{first}{rgb}{1.0, .83, 0.3}
\definecolor{second}{rgb}{1.0, 0.93, 0.7}
\def \first {\cellcolor{first}}
\def \second {\cellcolor{second}}
\def \third {}

\begin{table}[t]
\begin{center}

\setlength{\tabcolsep}{2pt}
\resizebox{\linewidth}{!}{
\begin{tabular}{l c c c c c c c c }
\toprule
        \multirow{2}{*}{Method} & \multicolumn{3}{c}{New Days} & \multicolumn{3}{c}{VISOR} \\   
        \cmidrule(r){2-4} \cmidrule(r){5-7} 
        & @0.05 & @0.1 & @0.15 & @0.05 & @0.1 & @0.15 \\
        \midrule
        FrankMocap \cite{frankmocap} & 16.1 & 41.4 & 60.2 & 16.8 & 45.6 & 66.2 \\
        METRO \cite{metro} & 14.7 & 38.8 & 57.3 & 16.8 & 45.4 & 65.7 \\
        MeshGraphormer \cite{meshgraphormer} & 16.8 & 42.0 & 59.7 & 19.1 & 48.5 & 67.4 \\
        HandOccNet (param)\cite{handoccnet} & 9.1 & 28.4 & 47.8 & 8.1 & 27.7 & 49.3\\
        HandOccNet (no param) \cite{handoccnet} & 13.7 & 39.1 & 59.3 & 12.4 & 38.7 & 61.8 \\
        HaMeR \cite{hamer} & \second {48.0} & \second {78.0} & \second {88.8} & \second{43.0} & \second{76.9} & \second{89.3} \\
        \ours~(Ours) & \first 48.6 &	\first  79.0 &	\first  89.9 & \first  46.8 &	\first  79.5 &	\first  90.9 \\
\bottomrule
\end{tabular}
}
\vspace{-.5em}
\caption{\textbf{Evaluation of single image pose.} We report PCK at different thresholds on benchmark HInt. We compare \ours~with multiple image-based hand pose estimation baselines.}
\label{tab:image-pose}
\vspace{-1em}
\end{center}
\end{table}

\definecolor{first}{rgb}{1.0, .83, 0.3}
\definecolor{second}{rgb}{1.0, 0.93, 0.7}
\def \first {\cellcolor{first}}
\def \second {\cellcolor{second}}
\def \third {}

\begin{table}[t]
\footnotesize
\begin{center}
\setlength{\tabcolsep}{2pt}
\resizebox{\linewidth}{!}{
\begin{tabular}{l c c c c c c }
\toprule
 & GA-MPJPE & FA-MPJPE & ACC-NORM & PA-MPJPE 
 \\
\midrule
No Image Data & 2.34 & 5.52 & 1.68 & 0.55 \\
No Cross-View SA & 2.82 & 6.63 & \first \textbf{1.41} & 0.65 \\
No Global CA  & 2.25 & \second 5.12 & 1.62 &  0.54 \\
Flip Order & \second 2.20 & 5.23 & 1.62 & \first \textbf{0.53} \\
Full Model & \first \textbf{1.83} & \first \textbf{4.27} & \second 1.59 & 0.54 \\
\bottomrule
\end{tabular}
}
\vspace{-.5em}
\caption{\textbf{Ablation studies.} We compare our full model with variants that do not mix training batch with image data, do not have one of the attention layers, or have attention layers with flipped order on one view of ARCTIC-EXO.}
\label{tab:ablation}
\vspace{-1.5em}
\end{center}

\end{table}

\begin{figure*}
    \centering
    \includegraphics[width=\linewidth]{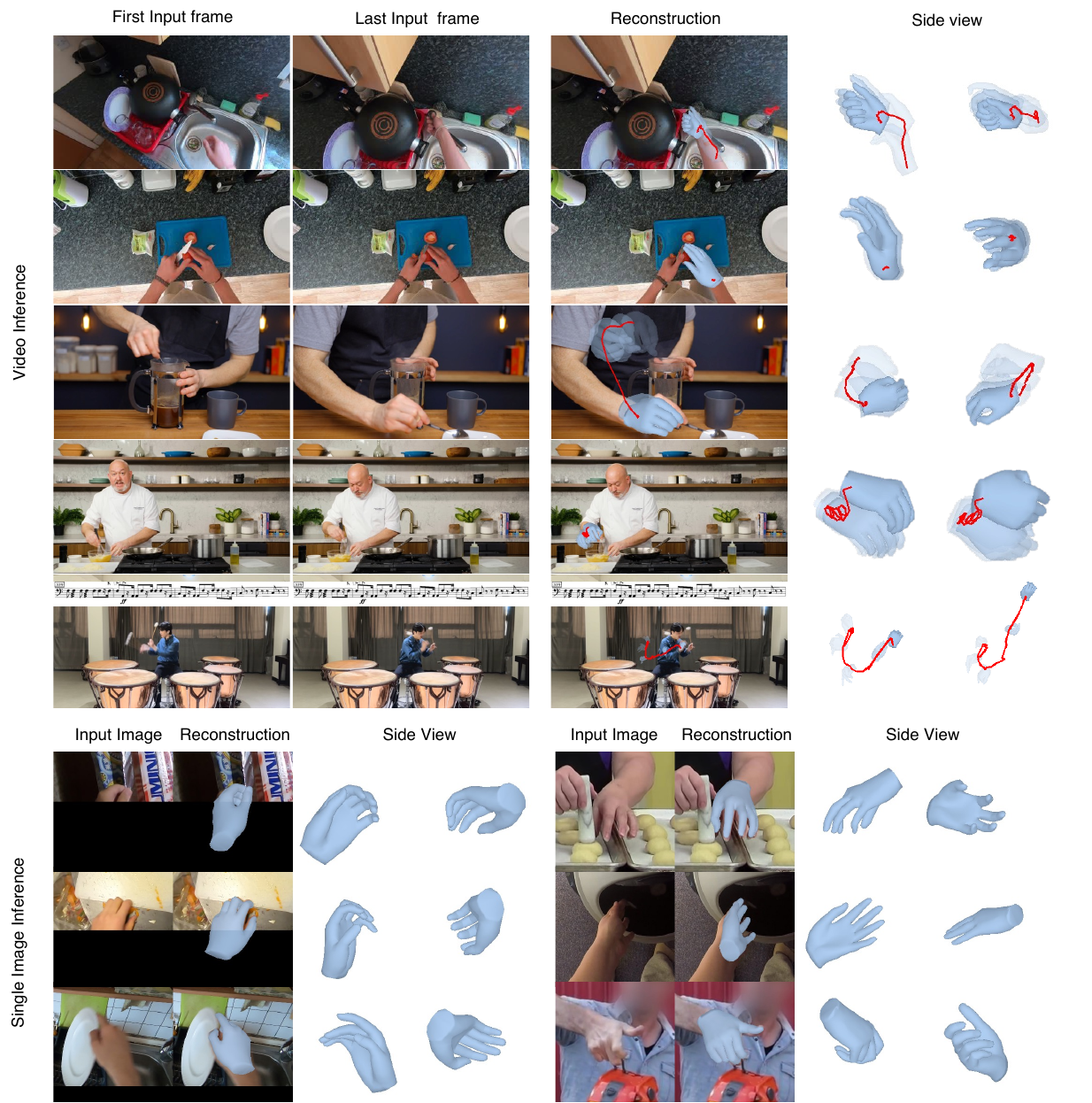}
    \caption{
    \textbf{Qualitative Results.} We show more qualitative results of our general model \ours. Upper: Given videos from various datasets or from the Internet, we show reconstructed hand pose and trajectory from the original image view and from two side views. Lower: Given single images from the HInt dataset or the Internet, we show reconstructed poses from the original image view and side views. }
    \label{fig:result}
\end{figure*}
\subsection{Ablation Study}
\label{sec:ablation}
In Table~\ref{tab:ablation}, we analyze the effect of our design choices by comparing variants of our models that train and test on one allocentric view in ARCTIC-EXO. When we only train on video datasets without auxiliary image datasets, the model does not generalize well to the test split probably because the appearance of video training data is not diverse enough. Without cross-view attention (No Cross-View SA), the global trajectory performs the worst as it cannot fuse information from adjacent frames.  Without attending to the global frame (No Global CA), the trajectory is less smoother (higher ACC-NORM) probably because the original frame, which varies less than hand crops in a sequence, stabilizes the predictions.   We also try flipping the order of the two attention layers in the transformer decoder and find that the current design leads to the optimal global trajectory.  

\subsection{In-the-Wild Results}
\label{sec:itw}
Finally, we show more qualitative results in Fig.~\ref{fig:result}, including predictions from in-the-wild videos (top) and single images (bottom). 
\ours can handle different hand sizes and different camera viewpoints. It can handle fast hand motions like drumming (row 5) while it correctly predicts the holding hand being static (row 2).  The reconstruction also includes challenging cyclic motions like whisking (row 4).  In the bottom row of Figure~\ref{fig:teaser}, the reconstructed hand trajectory desirably follows a straight line back and forth. 
The image reconstruction results show good 2D alignment of hand pose, even with severe hand truncation.  We encourage readers to see \supmat videos. 

\section{Conclusion}

In this work, we present \ours, the first feed-forward method to predict 4D hand trajectory in global coordinates. 
To address the issue of limited video training data, we adapt a high-quality image-based model with novel attention layers and train it on both video and auxiliary image datasets. The adaption is simple and effective, including two types of attention and an additional prediction head. 
While we have achieved state-of-the-art results, the method assumes a reliable 2D hand tracking system. An interesting direction would be jointly tracking and reconstructing. Another possible direction is to infer hand trajectory in the world instead of camera coordinates. We hope that \ours can serve as a useful tool for downstream tasks like hand-object interaction, robotics, and AR/VR. 

\paragraph{Acknolwedgement. }
The authors would like to thank Georgios Pavlakos, Dandan Shan, and Soyong Shin for comparisons with baselines HaMeR and WHAM. Yufei would like to appreciate Shashank Tripathi, Markos Diomataris, and Sai Kumar Dwivedi for fruitful discussions. Part of the work is done when Yufei was an intern at Max-Planck Institute. We also thank Ruihan Gao and Sheng-Yu Wang for proofread.

\noindent\textbf{Disclosure}
MJB has received research gift funds from Adobe, Intel, Nvidia, Meta/Facebook, and Amazon.  MJB has financial interests in Amazon and Meshcapade GmbH.  While MJB is a co-founder and Chief Scientist at Meshcapade, his research in this project was performed solely at, and funded solely by, the Max Planck Society. Yufei’s PhD research is partially supported by a Google Gift.

{
    \small
    \bibliographystyle{ieeenat_fullname}
    \bibliography{main_arxiv}
}

\clearpage
\setcounter{page}{1}
\setcounter{table}{4}
\setcounter{figure}{6}
\maketitlesupplementary
\appendix

In supplementary material, we provide additional quantitative comparisons mentioned in the main paper (Sec.~\ref{sec:more-image}. 
We also provide further details on implementing network and evaluation metrics (Sec.~\ref{sec:detail}). Finally, we visualize more comparisons and results in supplementary videos. 

\section{Image-based Hand Pose Estimation}
\label{sec:more-image}
As mentioned in Section 4.2, we report quantitative comparisons between \ours~general model and other image-based baselines. Table~\ref{tab:image-pose-ho3d} reports 3D pose evaluation and  Table~\ref{tab:image-pose-hint} reports their 2D projections. 

\paragraph{All Splits on HInt}
\ours ~outperforms all other baselines in terms of 2D pose alignment on the more appearance-diverse benchmark HInt. The advantage is even more significant on the occluded split. This is likely because our global spatial attention (Global CA) to the original frame provides more context under occlusion.

\definecolor{first}{rgb}{1.0, .83, 0.3}
\definecolor{second}{rgb}{1.0, 0.93, 0.7}
\def \first {\cellcolor{first}}
\def \second {\cellcolor{second}}
\def \third {}

\begin{table}[t]
\begin{center}

\setlength{\tabcolsep}{2pt}
\resizebox{\linewidth}{!}{
\begin{tabular}{llcccccc}
\hline
& {Method}                & \multicolumn{3}{c}{{New Days}} & \multicolumn{3}{c}{{VISOR}} \\
&                               & @0.05  & @0.1   & @0.15  & @0.05  & @0.1   & @0.15  \\ \hline
\multirow{7}{*}{\rotatebox[origin=c]{90}{All Joints}}& FrankMocap\cite{frankmocap}                     & 16.1   & 41.4   & 60.2   & 16.8   & 45.6   & 66.2   \\
& METRO\cite{metro}                          & 14.7   & 38.8   & 57.3   & 16.8   & 45.4   & 65.7   \\
& Mesh Graphormer \cite{meshgraphormer}               & 16.8   & 42.0   & 59.7   & 19.1   & 48.5   & 67.4   \\
& HandOccNet (param) \cite{handoccnet}             & 9.1    & 28.4   & 47.8   & 8.1    & 27.7   & 49.3   \\
& HandOccNet (no param)   \cite{handoccnet}       & 13.7   & 39.1   & 59.3   & 12.4   & 38.7   & 61.8   \\
& HaMeR    \cite{hamer}                      & \second 48.0   & \second 78.0   & \second88.8   & \second43.0   & \second76.9   & \second 89.3   \\
& \ours (Ours)                   & \first{48.6}   & \first{79.0}   & \first{89.9}   & \first{46.8}   & \first{79.5}   & \first{90.9}   \\ \hline                    
\multirow{7}{*}{\rotatebox[origin=c]{90}{Visible Joints}} & FrankMocap                     & 20.1   & 49.2   & 67.6   & 20.4   & 52.3   & 71.6   \\
& METRO                          & 19.2   & 47.6   & 66.0   & 19.7   & 51.9   & 72.0   \\
& Mesh Graphormer                & 22.3   & 51.6   & 68.8   & 23.6   & 56.4   & 74.7   \\
& HandOccNet (param)             & 10.2   & 31.4   & 51.2   & 8.5    & 27.9   & 49.8   \\
& HandOccNet (no param)          & 15.7   & 43.4   & 64.0   & 13.1   & 39.9   & 63.2   \\
& HaMeR                          & \second 60.8   & \second 87.9   & \second 94.4   & \second 56.6   & \second 88.0   & \second 94.7   \\
& \ours (Ours)                   & \first{60.6}   & \first{88.5}   & \first{95.0}   & \first{60.7}   & \first{89.3}   & \first{95.5}   \\ \hline
\multirow{7}{*}{\rotatebox[origin=c]{90}{Occluded Joints}} &   FrankMocap                     & 9.2    & 28.0   & 46.9   & 11.0   & 33.0   & 55.0   \\
& METRO                          & 7.0    & 23.6   & 42.4   & 10.2   & 32.4   & 53.9   \\
& Mesh Graphormer                & 7.9    & 25.7   & 44.3   & 10.9   & 33.3   & 54.1   \\
& HandOccNet (param)             & 7.2    & 23.5   & 42.4   & 7.4    & 26.1   & 46.7   \\
& HandOccNet (no param)          & 9.8    & 31.2   & 50.8   & 9.9    & 33.7   & 55.4   \\
& HaMeR                          & \second 27.2   & \second 60.8   & \second 78.9   & \second 25.9   & \second 60.8   & \second 80.7   \\
& \ours (Ours)                   & \first{29.4}   & \first{62.2}   & \first{80.4}   & \first{29.7}   & \first{64.6}   & \first{83.1}   \\ \hline
\end{tabular}%

}
\caption{\first{Evaluation of single image pose.} We report PCK at different thresholds on benchmark HInt~\cite{hamer}. We compare \ours~with multiple image-based hand pose estimation baselines.}
\label{tab:image-pose-hint}
\vspace{-1em}
\end{center}
\end{table}

\definecolor{first}{rgb}{1.0, .83, 0.3}
\definecolor{second}{rgb}{1.0, 0.93, 0.7}
\def \first {\cellcolor{first}}
\def \second {\cellcolor{second}}
\def \third {}

\begin{table}[t]
\begin{center}

\setlength{\tabcolsep}{2pt}
\resizebox{\linewidth}{!}{

\begin{tabular}{lcccccc}
\hline
\textbf{Method}         & \textbf{AUC\textsubscript{J} ↑} & \textbf{PA-MPJPE ↓} & \textbf{AUC\textsubscript{V} ↑} & \textbf{PA-MPVPE ↓} & \textbf{F@5 ↑} & \textbf{F@15 ↑} \\ \hline
Liu et al.    \cite{liu2021semi}          & 0.803                           & 9.9                 & 0.810                           & 9.5                 & 0.528         & 0.956          \\
HandOccNet   \cite{handoccnet}           & 0.819                           & 9.1                 & 0.819                           & 8.8                 & 0.564         & 0.963          \\
I2UV-HandNet \cite{chen2021i2uv}            & 0.804                           & 9.9                 & 0.799                           & 10.1                & 0.500         & 0.943          \\
Hampali et al. \cite{ho3d}        & 0.788                           & 10.7                & 0.790                           & 10.6                & 0.506         & 0.942          \\
Hasson et al. \cite{hasson2019learning}          & 0.780                           & 11.2                & 0.777                           & 11.1                & 0.464         & 0.939          \\
ArtiBoost     \cite{yang2022artiboost}          & 0.773                           & 11.4                & 0.782                           & 11.4                & 0.488         & 0.944          \\
Pose2Mesh    \cite{choi2020pose2mesh}           & 0.754                           & 12.5                & 0.749                           & 12.7                & 0.441         & 0.909          \\
I2L-MeshNet   \cite{moon2020i2l}          & 0.775                           & 11.2                & 0.722                           & 13.9                & 0.409         & 0.932          \\
METRO     \cite{metro}              & 0.792                           & 10.4                & 0.779                           & 11.1                & 0.484         & 0.946          \\
MobRecon    \cite{chen2022mobrecon}            & -                               & 9.2                 & -                               & 9.4                 & 0.538         & 0.957          \\
Keypoint Trans  \cite{hampali2022keypoint}        & 0.786                           & 10.8                & -                               & -                   & -             & -              \\
AMVUR       \cite{jiang2023probabilistic}            & {0.835}                  & {8.3}        & {0.836}                  & {8.2}        & {0.608} & 0.965          \\
HaMeR      \cite{hamer}             & \first{0.846}                   & \first{7.7}         & \first{0.841}                   & \first{7.9}         & \first{0.635}  & \first{0.980}  \\
\ours (Ours)            & \second 0.842                           &  \second 8.0                 & \second 0.839                           & \second 8.1                 & \second 0.628         & \first{0.980}  \\ \hline
\end{tabular}
}
\caption{\first{Evaluation of single image pose.} We report  quality of 3D hand pose on benchmark HO3D. We compare \ours~with multiple image-based hand pose estimation baselines.}
\label{tab:image-pose-ho3d}
\vspace{-1em}
\end{center}
\end{table}

\paragraph{3D Hand Pose Benchmark}
\ours~performs comparably in terms of 3D hand poses on HO3D (second best across all methods with gaps to the best model less than 1\%). It is slightly worse than HaMeR probably because the HO3D dataset is sampled less in the combination of video datasets. Yet, \ours~predicts much more realistic 4D trajectories.

\section{Implementation Details}
\label{sec:detail}
\paragraph{Network Implementation Details}
We use AdamW optimizer with learning rate $1e-4$. We train on eight H100 with batch size 16 (8 for video and 8 for images) for 1000000 iterations. We use the same weights as that in HaMeR to mix data from different image datasets. The (unnormalized) sampled weights to mix video datasets are 0.25 (ARCTIC-EGO), 0.15 (ARCTIC-EXO), 0.05 (DexYCB), 0.05 (H2O), 0.05 (HO3D), 0.15 (InterHand2.6M). 

In test time optimization, we use AdamW optimization with learning rate $1e-3$ and optimize for 1000 iterations for each sequence. The optimization objective consists of 2D reprojection loss with ground truth 2D keypoints, acceleration loss of predicted 3D and 2D keypoints, and distance to the original predictions.

\paragraph{Evaluation Metrics. }
To calculate GA/FA-MPJPE, we align the predicted trajectory with the ground truth by searching for an affine transformation (isometric scale, rotation, and translation) between selected keypoints. In GA-MPJPE, the selected keypoints are all joints from all frames while in FA-MPJPE, they are all joints from the first frame. 

In ARCTIC, each video takes about 30 seconds. Following the literature on whole-body reconstruction in global coordinates~\cite{wham,slahmr}, we clip long sequences into shorter clips of the same length (60) before computing global trajectory metrics. 

\section{More Results in Videos}
\label{sec:video}

We provide more results in Supplementary Videos, including comparisons with other models (Fig.~4), comparisons before and after optimization (Fig.~5), and more qualitative results from \ours (Fig.~1 and 6).

\end{document}